\documentclass[10pt,twocolumn,letterpaper]{article}

\usepackage{iccv}              

\usepackage{xcolor}
\usepackage{caption}
\usepackage{enumitem}
\usepackage{tcolorbox}

\newlist{promptlist}{itemize}{3}
\setlist[promptlist]{
  label=\textbullet,
  leftmargin=*,
  topsep=2pt,
  partopsep=0pt,
  parsep=2pt,
  font=\ttfamily\small
}

\newlist{promptenum}{enumerate}{3}
\setlist[promptenum]{
  label=\arabic*.,
  leftmargin=*,
  topsep=2pt,
  partopsep=0pt,
  parsep=2pt,
  font=\ttfamily\small
}

\newtcolorbox{promptbox}{
  colback=gray!10,
  colframe=black!80,
  boxrule=0.5pt,
  arc=0mm,
  fontupper=\ttfamily\small,
  boxsep=2mm,
}

\definecolor{iccvblue}{rgb}{0.21,0.49,0.74}
\usepackage[pagebackref,breaklinks,colorlinks,allcolors=iccvblue]{hyperref}

\makeatletter
\renewcommand*{\@fnsymbol}[1]{\ensuremath{\ifcase#1\or *\or \dagger\or \ddagger\or \mathsection\or \mathparagraph\or \|\or **\or \dagger\dagger\or \ddagger\ddagger \else\@ctrerr\fi}}
\makeatother

\def\paperID{9} 
\def\confName{ICCV}
\def\confYear{2025}

\title{EgoInstruct: An Egocentric Video Dataset of Face-to-face Instructional Interactions with Multi-modal LLM Benchmarking}

\author{Yuki Sakai\thanks{Co-first authors}, Ryosuke Furuta\footnotemark[1], Juichun Yen, and Yoichi Sato\\
The University of Tokyo\\
{\tt\small \{sakai-y, furuta, yen, ysato\}@iis.u-tokyo.ac.jp}
}

\begin{document}
\maketitle

\begin{abstract}
  Analyzing instructional interactions between an instructor and a learner who are co-present in the same physical space is a critical problem for educational support and skill transfer. 
  Yet such face-to-face instructional scenes have not been systematically studied in computer vision. 
  We identify two key reasons: i) the lack of suitable datasets and ii) limited analytical techniques.

  To address this gap, we present a new egocentric video dataset of face-to-face instruction and provide ground-truth annotations for two fundamental tasks that serve as a first step toward a comprehensive understanding of instructional interactions: procedural step segmentation and conversation-state classification. 
  Using this dataset, we benchmark multimodal large language models (MLLMs) against conventional task-specific models. 
  Since face-to-face instruction involves multiple modalities (speech content and prosody, gaze and body motion, and visual context), effective understanding requires methods that handle verbal and nonverbal communication in an integrated manner. 
  Accordingly, we evaluate recently introduced MLLMs that jointly process images, audio, and text. 
  This evaluation quantifies the extent to which current machine learning models understand face-to-face instructional scenes. 
  In experiments, MLLMs outperform specialized baselines even without task-specific fine-tuning, suggesting their promise for holistic understanding of instructional interactions.
\end{abstract}

\section{Introduction}
\label{sec:intro}

\begin{figure}[t]
  \centering
      \centering
      \includegraphics[width=0.75\linewidth]{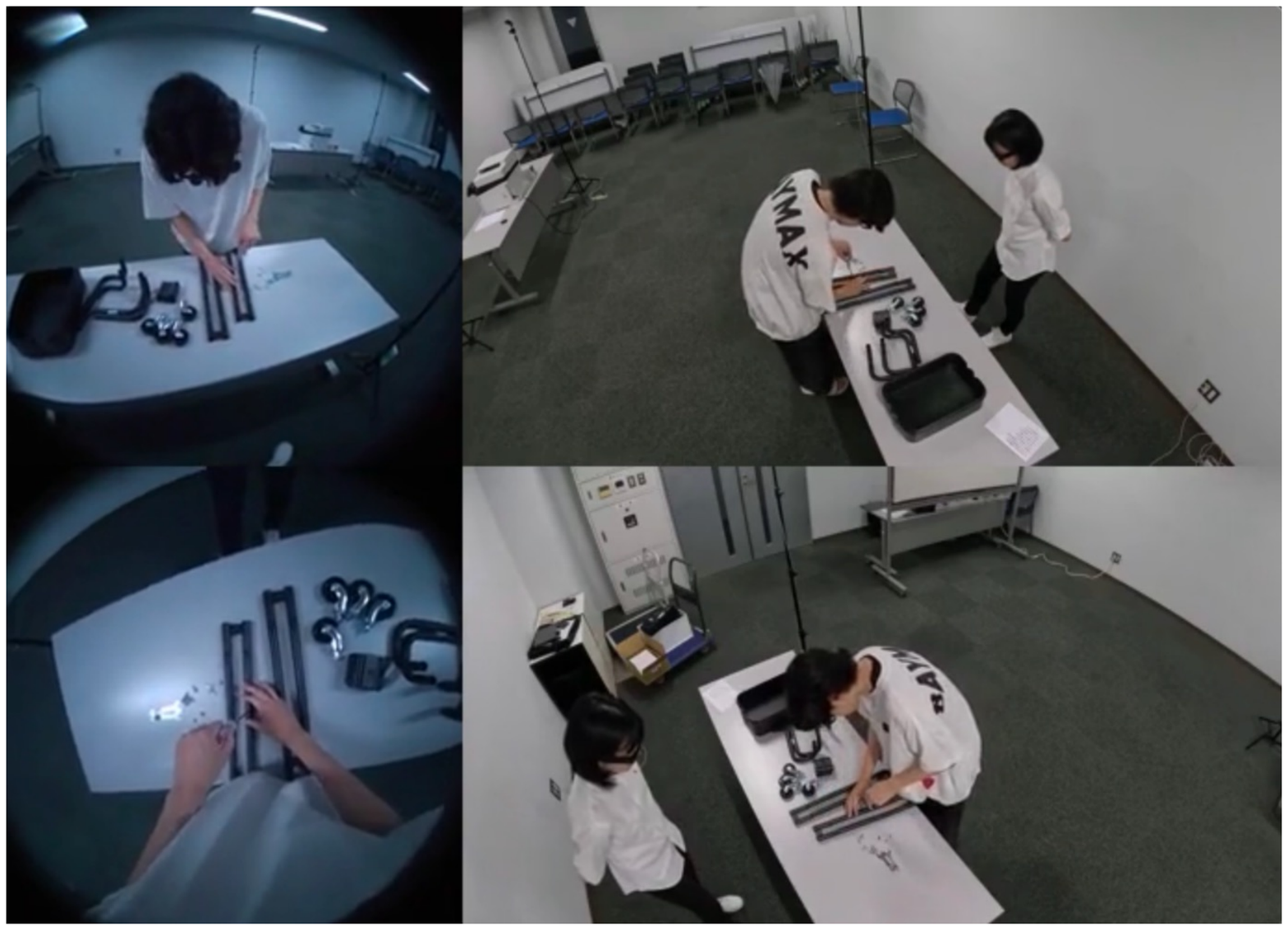}
      \caption{A sample face-to-face instructional scene between an instructor and a learner.}
      \label{fig:KibanA-overview}
  \end{figure}

  With an aging expert workforce, industry increasingly demands effective training of younger technicians. Tacit knowledge and practical know-how accumulated over years are often difficult to communicate fully through documents and manuals alone. Some aspects have traditionally been conveyed only through direct instruction, where an expert instructor interacts with a novice learner in the same space. Consequently, analyzing face-to-face instructional scenes using computer vision is an important challenge for educational support and skill transfer.

  Against this backdrop, egocentric video has attracted considerable attention. Compared to third-person footage, which captures full-body pose and the surrounding environment, egocentric video records the scene from the camera wearer’s viewpoint. Analyzing egocentric videos that include natural conversation, nods, and hand gestures offers the potential to reveal intentions and motivations underlying human behavior. Because egocentric videos are recorded with head- or body-mounted wearable cameras, they inherently encode the wearer’s viewpoint and environment, as well as cues to interest and attention. This makes it possible to capture crucial information such as expert gaze patterns and attentional strategies. Even as experts retire, egocentric recordings can preserve rich information; analyzing such videos may also allow novices to virtually experience an expert’s perspective and deepen their understanding during training.
  
  While many studies have tackled action and procedure understanding from egocentric video \cite{nunez2022egocentric}, most do not target face-to-face instructional scenes. In instructional settings, beyond verbal directions from the instructor, there are diverse interactions: nonverbal communication including gestures, learners’ backchanneling and questions, and more. Understanding these interactional elements holistically is essential for improving training effectiveness. However, modeling interaction is difficult since human behavior changes fluidly through mutual influence. Natural interactions that include conversation and bodily motion rarely conform to a single consistent template. Learning from data using machine learning is therefore appealing, yet current research on interaction analysis faces two issues: i) a shortage of datasets specifically capturing face-to-face scenes and ii) limited techniques for analyzing face-to-face instructional scenes.
  
  There are many egocentric datasets from a worker’s viewpoint \cite{ego4d,grauman2024ego,epickitchen}, but most focus on single-person tasks and lack genuine face-to-face interaction. Although the number of egocentric datasets is increasing, large-scale multimodal datasets that capture time-synchronized egocentric views from multiple people engaged in co-present interactions are scarce. In particular, we are not aware of datasets that focus on instruction or collaborative task execution.
  
  We therefore present a new egocentric dataset targeting face-to-face instructional scenes (Fig. \ref{fig:KibanA-overview}). We collected 38 sessions totaling 8 hours in which an instructor teaches tasks such as furniture assembly or replacing a camera battery while a learner executes the task. All participants wore Aria Glasses \cite{ariaglasses}, which provide, in addition to first-person video, gaze information, skeletal keypoints for the wrist and palm, and head pose estimated from SLAM-based camera localization. Having synchronized egocentric views and these signals from both the instructor and the learner is a distinctive feature compared with prior work. Although we do not analyze gaze or head pose in this paper, we expect gaze (which indicates where the instructor and learner look while acting) and relative 3D positions obtained by SLAM to serve as important cues for future interaction analysis.
  
  Building on this dataset, we define two fundamental tasks as the first step toward a holistic understanding of instructional interactions. The first is procedural step segmentation, which aims to segment task progress into steps from both the instructor’s and the learner’s viewpoints, thereby extracting procedural information useful for skill transfer. The second is conversation-state classification of utterances, for example, “the learner asks the instructor a question” or “the instructor gives instructions.” Classifying conversation states structures the interaction flow and helps analyze instruction; it can surface difficult points and important tricks while enabling the learner to gradually acquire skills, including tacit knowledge.
  
  For our experiments, we annotate procedural steps from both viewpoints, and we also annotate the utterance spans, their transcripts, and the conversation state (e.g., “question from learner,” “instruction from instructor”). Our goal is to provide a foundation for the analysis of face-to-face instructional scenes.
  
  To understand behavior, content, and meaning in such scenes, methods must integrate verbal and nonverbal communication because multiple modalities (language, speech, and vision) are involved simultaneously. Recently, multimodal large language models (MLLMs) that jointly process images, audio, and text have emerged, offering a new avenue for this challenge and enabling more holistic scene understanding. Using our dataset, we therefore evaluate MLLMs alongside conventional specialized models to clarify how well current machine learning models understand face-to-face instruction. In our experiments, MLLMs outperform task-specific models without additional fine-tuning, indicating their potential for comprehensive understanding of face-to-face instructional scenes.
  
  Our contributions are summarized as follows:
  \begin{itemize}
  \item To our knowledge, this is the first study that directly targets understanding of face-to-face instructional scenes.
  \item We construct a new egocentric dataset of face-to-face instructional scenes. We define two tasks (procedural step segmentation and conversation-state classification) and provide ground-truth annotations.
  \item We benchmark MLLMs against specialized baselines on both tasks. 
  \end{itemize}

\section{Related work}

\subsection{Egocentric Video Datasets}

\begin{table*}[t]
  \centering
  \caption{Comparison with related datasets.}
  \setlength{\tabcolsep}{2mm}{
  \begin{tabular}{@{}lcccc@{}}
    \toprule
    Dataset & Has human interaction & Setting & Total duration [h] \\
    \midrule
    Ego4D \cite{ego4d} &  & Daily tasks & 3,670\\
    EgoExo4D \cite{grauman2024ego} & & Skilled activity & 1,286 \\
    Epic-Kitchen-100 \cite{epickitchen} &  & Cooking & 100\\
    Assembly101 \cite{assembly101} &  & Toy assembly & 167\\
    Holoassist \cite{holoassist} &  & Instruction & 166\\
    EgoCom \cite{egocom} & \checkmark & Group conversation & 38.5\\
    AEA \cite{AEA} & \checkmark & Daily tasks & 7.3\\ \hline
    \textbf{EgoInstruct (ours)} & \checkmark & Instruction & 8\\
    \bottomrule
  \end{tabular}
  \label{tab:diff-dataset}}
\end{table*}

A large number of egocentric datasets have been proposed \cite{ego4d,epickitchen,assembly101}. EPIC-KITCHENS \cite{epickitchen} focuses on activities in the kitchen, and Ego4D \cite{ego4d} provides a large-scale egocentric benchmark with more than 3,670 hours of everyday activities. Ego-Exo4D \cite{grauman2024ego} captures skilled activities with time-synchronized first- and third-person views. However, these datasets mainly focus on single-person tasks and do not target multi-person interaction; they typically include egocentric views only from the task performer.

To analyze tasks and interactions involving multiple people, time-synchronized multi-view datasets are essential. Egocom \cite{egocom} captures natural conversations in activities like high-fives and card games, providing egocentric footage along with speech recognition annotations and benchmarks for conversation turn prediction. The Aria Everyday Activities dataset \cite{AEA} collects daily-life tasks such as making coffee or tidying a room, including tasks performed solo or by two people. Although several synchronized multi-person datasets exist, none specifically focuses on face-to-face instruction. Our dataset is designed to fill this gap (Table \ref{tab:diff-dataset}).

\subsection{Video Understanding and Procedure Recognition}
Video understanding encompasses downstream tasks such as procedure recognition, step recognition, and step prediction. Among large-scale instructional corpora, HowTo100M \cite{howto100m} collects YouTube how-to videos, and subsequent work builds procedure knowledge graphs to infer the structured organization of tasks demonstrated in videos.

Recently, many studies leverage large language models for video understanding \cite{PREGO,MM-VID,longVML,video-streaming}. MM-VID \cite{MM-VID} aggregates pre-processing signals such as scene detection and automatic speech recognition together with external video metadata, generates detailed textual descriptions, and then uses GPT-4 to produce scripts, including video summaries, based on those signals and clip-level features. LongVML \cite{video-streaming} segments long videos into shorter clips and combines local clip-level features with global semantics, which mitigates the loss of fine-grained information in long-form video understanding.

There are also methods that use egocentric video to discover object-use procedures. For example, Damen et al. \cite{you-do-i-learn} proposed a method to discover how objects are used from first-person recordings of predefined tasks. They leverage gaze to locate task-relevant objects, characterize their appearance and usage, and retrieve video snippets to explain how a recognized object was used previously. Although Paprika \cite{paprika} and You-Do, I-Learn \cite{you-do-i-learn} focus on procedure discovery, they assume single-actor task execution rather than interactive scenes. In contrast, we focus on the conversations that occur while an instructor teaches a learner.

\subsection{Action Segmentation}
Action segmentation temporally partitions a video by predicting a class label for each frame. As there is already a large body of work on this task \cite{ding2023temporal}, we briefly review representative methods. Lea et al. \cite{actionseg-1} introduced hierarchical temporal convolutions for efficient computation. MS-TCN \cite{MS-TCN} extends this line with dilated temporal convolutions, enabling long-range modeling at high temporal resolution. Building on the observation that action segmentation is closely related to language tasks, Yi et al. proposed a Transformer-based model, ASFormer \cite{asformer}.

Temporal segmentation models such as ASFormer have improved the detection of procedural steps and action intervals. However, most prior work targets single-person activities and has not been applied to scenarios with rich interaction such as face-to-face instruction. In this work, we use these methods together with MLLMs to analyze procedures and actions that unfold between an instructor and a learner from egocentric video.

\subsection{Multi-Person Interaction Analysis}
In multi-person scenes, prior work has examined attended objects \cite{ego-vision}, subtle bodily cues \cite{yonetani}, and the prediction of conversational turn-taking \cite{egocom}. Huang et al. studied jointly attended targets shared among multiple people. Yonetani et al. \cite{yonetani} proposed recognizing micro-actions and reactions from egocentric views, such as slight nods and small hand movements, suggesting that fine-grained social signals can be captured.

Research has also focused on speech in interactive settings \cite{holoassist,interaction-templete-2,interaction-templete}. HoloAssist \cite{holoassist} categorizes utterances and assigns labels within the dataset. Sauppé et al. \cite{interaction-templete-2} proposed conversational design patterns for human-robot interaction and, in doing so, observed human-to-human interactions across scenarios, identifying seven recurring patterns including question-answer pairs. Porfirio et al. \cite{interaction-templete} further defined utterance states within seven patterns such as conversation, collaboration, instruction, interview, and storytelling. Our definition of the conversation state is partly based on this instructional template.

The most closely related work to ours is HoloAssist \cite{holoassist}. It introduces a large-scale egocentric human interaction dataset in which two people collaboratively complete physical manipulation tasks. Their setting features remote instruction and is motivated by the goal of building an AI assistant that intervenes only when the performer makes a mistake. As a result, the instructor does not intervene proactively but provides help only after errors occur. In contrast, we focus on face-to-face instructional scenes in which the instructor teaches the task step by step and engages in proactive nonverbal communication. Accordingly, our dataset exhibits distinct characteristics relative to HoloAssist, particularly in setting, instructional goals, and interaction dynamics.

As discussed above, many egocentric datasets have been released and there is extensive research on understanding human behavior. There is also growing work on interaction scenes that focuses on attended objects, subtle movements, and utterance content. However, few egocentric datasets capture multiple people simultaneously. In particular, to our knowledge there is no egocentric dataset for instructional scenes that records both the instructor’s and the learner’s viewpoints for the purpose of completing a task. Moreover, little prior work examines both linguistic and non-linguistic directives in face-to-face instruction.

Our goal is to capture the details of interaction in face-to-face instructional scenes by constructing a new dataset and advancing interaction analysis through conversation-state classification and procedural step segmentation.

\section{Dataset Construction}

\begin{table*}[t]
  \centering
  \caption{Task-wise dataset statistics for EgoInstruct.}
  \setlength{\tabcolsep}{2mm}{
  \begin{tabular}{@{}lcccc@{}}
    \toprule
     Task & Positional relationship & \# steps & \# sessions & Average duration \\
    \midrule
    Cart assembly & face-to-face & 7 & 10 & 11m21s \\
    Chair assembly & face-to-face & 3 & 10 & 3m49s \\
    GoPro battery replacement & face-to-face & 6 & 10 & 3m3s  \\
    Printer toner replacement & side by side & 5 & 8 & 5m41s \\
    \bottomrule
  \end{tabular}
  \label{tab:task-data}}
\end{table*}

\begin{table}[t]
  \centering
  \caption{Recording devices and settings.}
  \setlength{\tabcolsep}{2mm}{
  \begin{tabular}{@{}lcccc@{}}
    \toprule
     Device & \# cameras & View & Resolution & FPS \\
    \midrule
    Aria Glass & 2 & FPV & 1400x1400 & 30 \\
    GoPro HERO11 & 2 & TPV & 4K (Wide) & 60\\
    \bottomrule
  \end{tabular}
  \label{tab:camera-data}}
\end{table}

\begin{figure}[t]
    \centering
    \includegraphics[width=\linewidth]{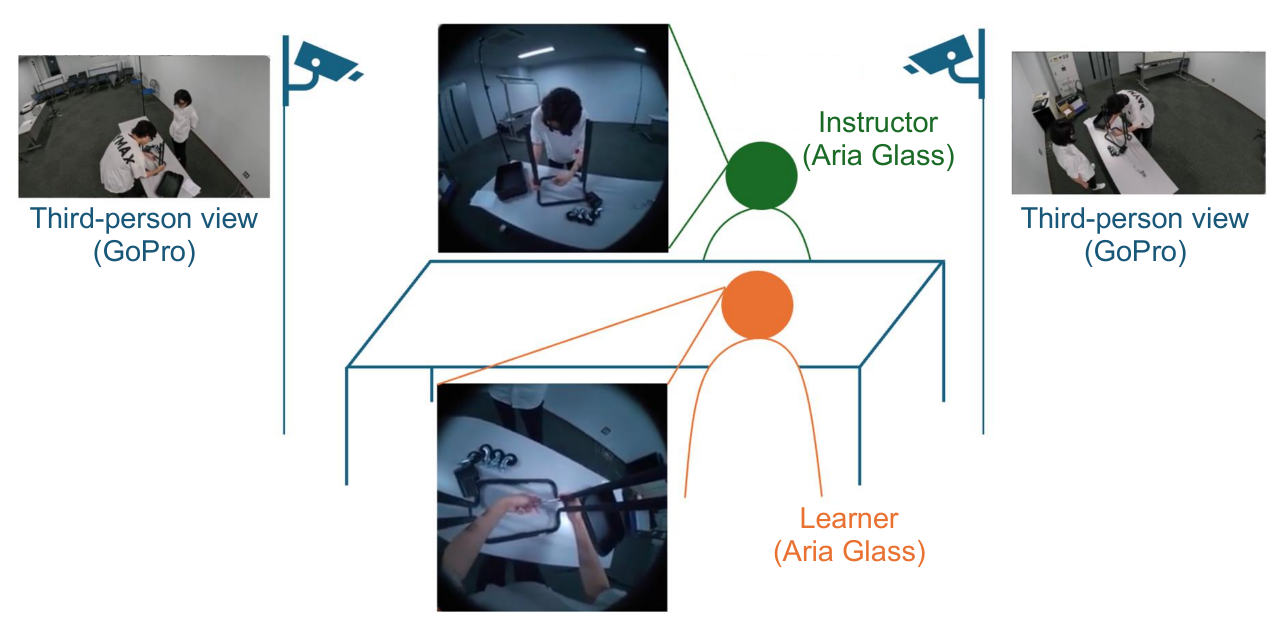}
    \caption{Positional relationship between the instructor and the learner, and sample frames from each camera.}
    \label{fig:KibanA-captured-position}
\end{figure}

\begin{figure*}[t]
    \begin{tabular}{ccc}
      \begin{minipage}[t]{0.30\linewidth}
        \centering
        \includegraphics[width=\textwidth]{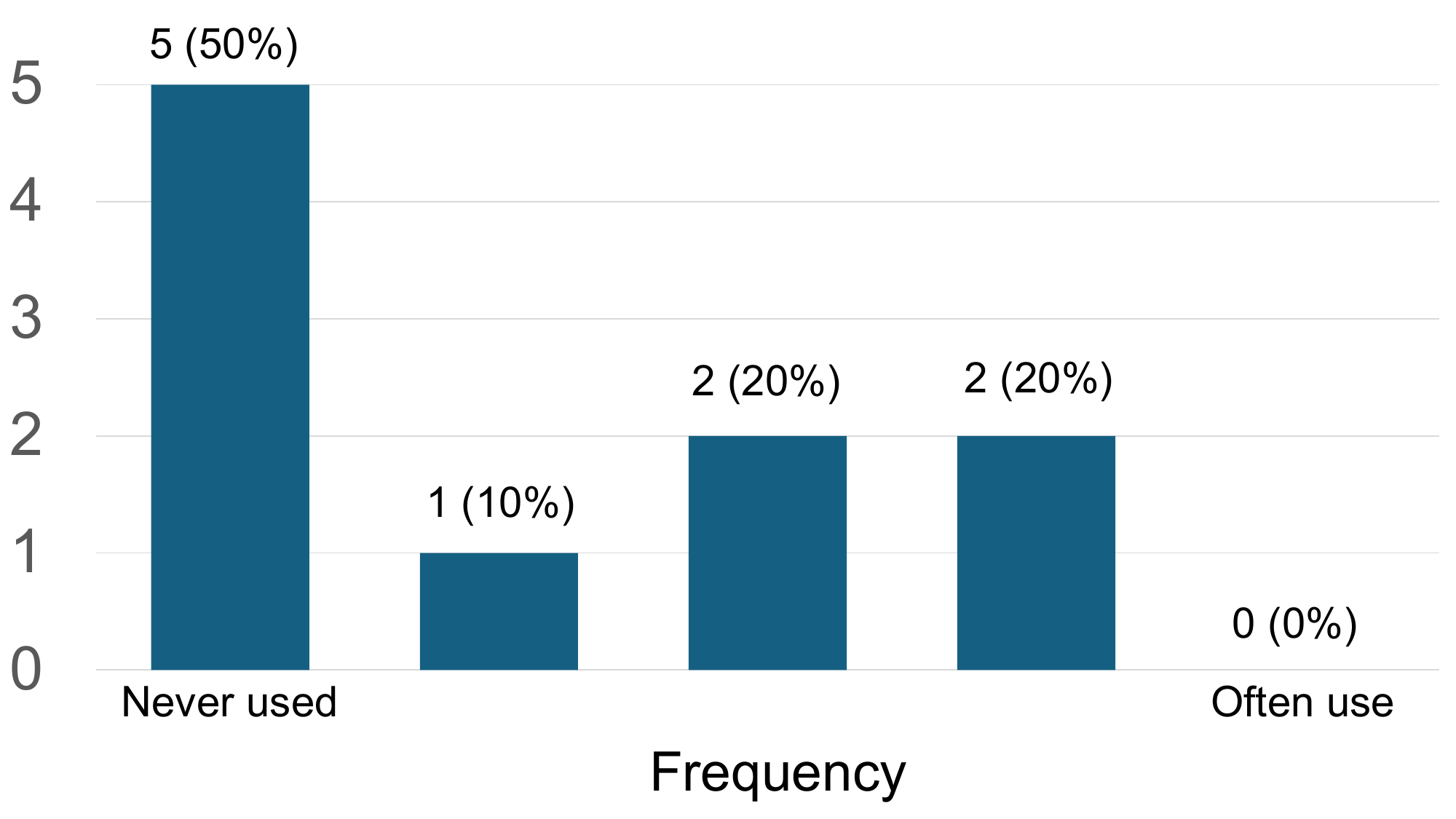}
        \subcaption{GoPro use}
        \label{composite}
      \end{minipage} &
      \begin{minipage}[t]{0.30\linewidth}
        \centering
        \includegraphics[width=\textwidth]{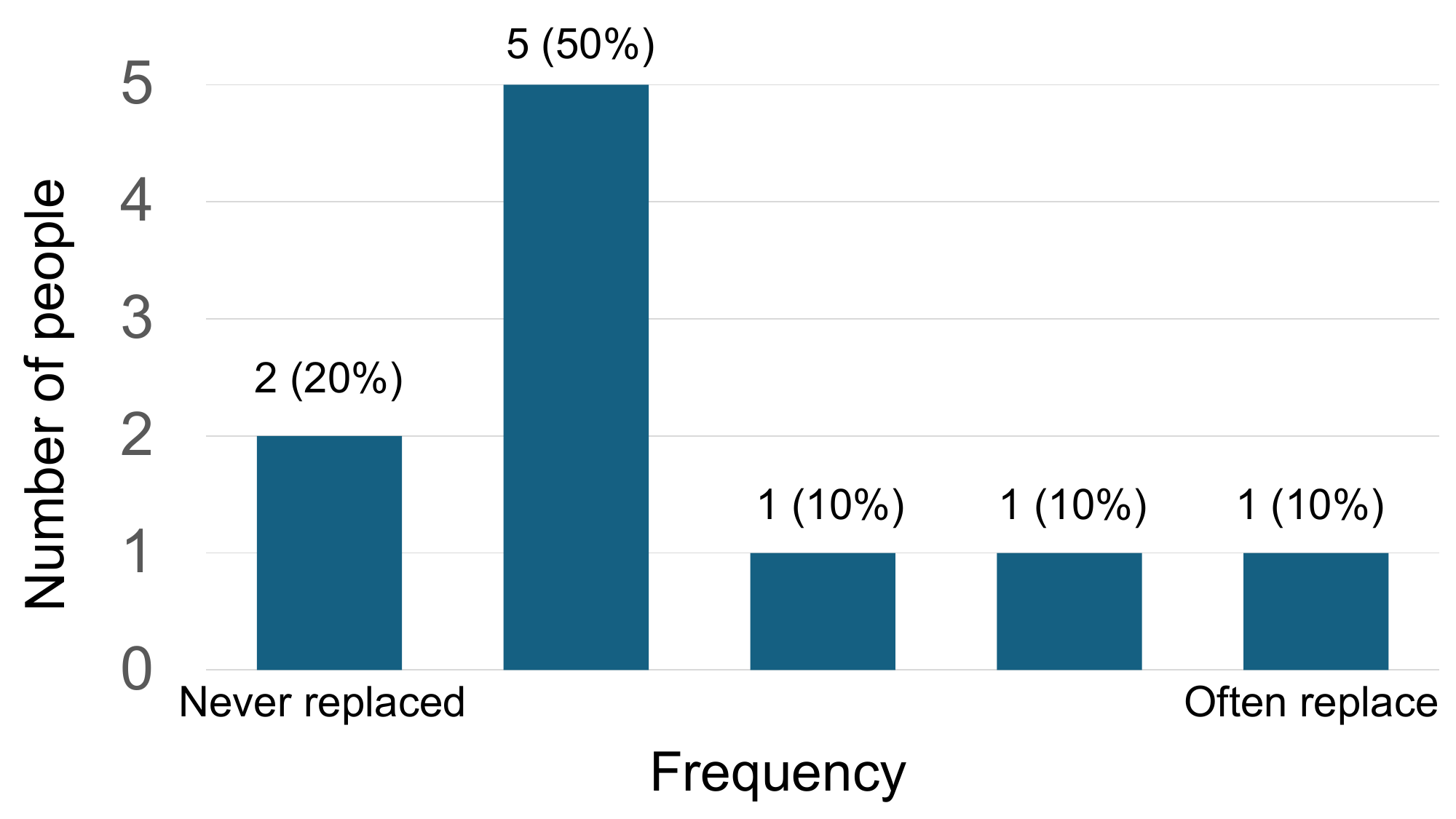}
        \subcaption{Printer toner replacement}
        \label{Gradation}
      \end{minipage} 
      \begin{minipage}[t]{0.30\linewidth}
        \centering
        \includegraphics[width=\textwidth]{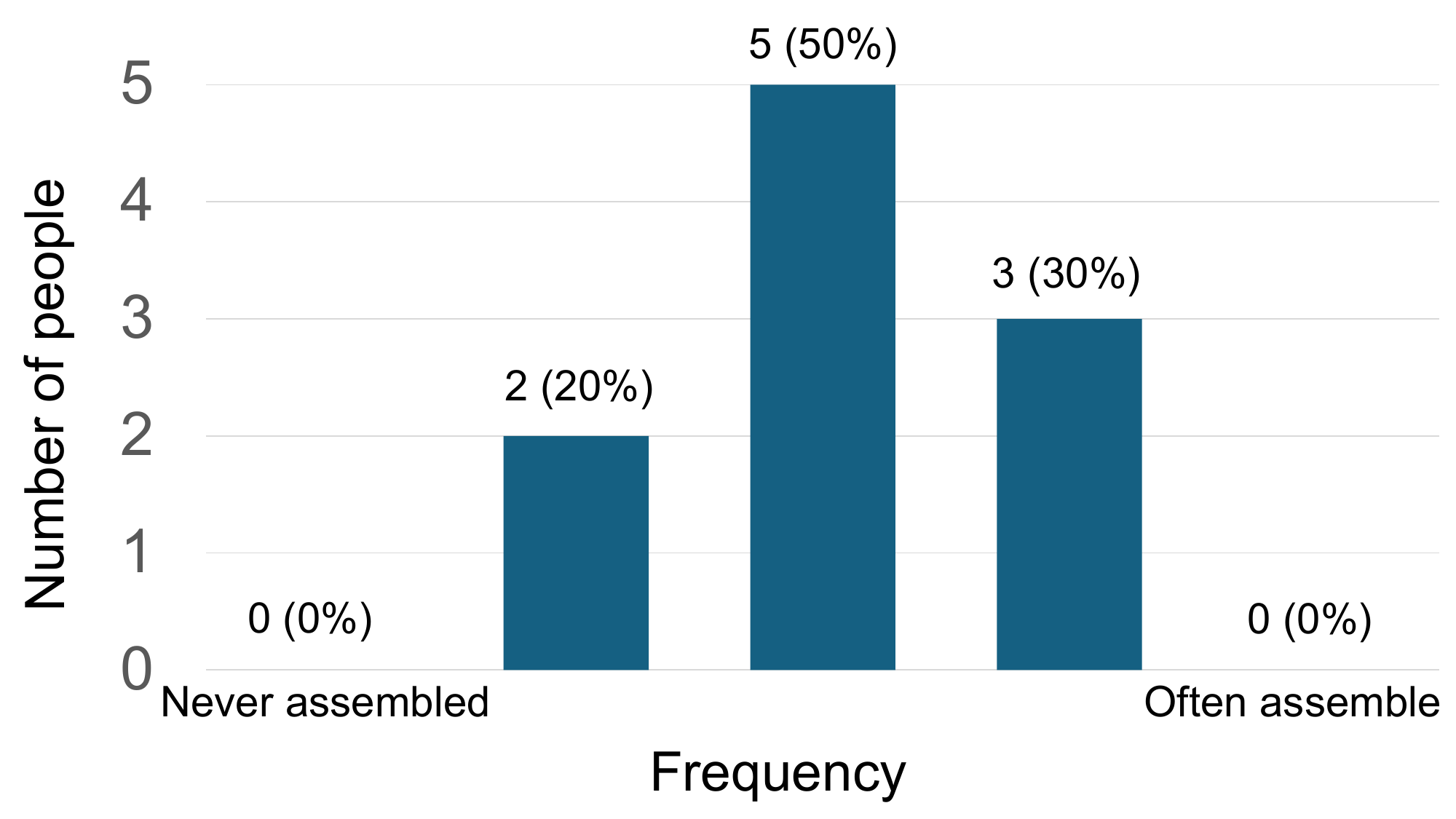}
        \subcaption{Furniture assembly}
        \label{fill}
      \end{minipage} &
    \end{tabular}
     \caption{Prior experience levels of the learners.}
     \label{fig:skill-level}
  \end{figure*}

Most existing egocentric datasets provide only the task performer’s viewpoint for single-person tasks. In contrast, we focus on the interaction between two participants by recording both the instructor and the learner as they face each other while the learner performs the task under instruction. To support interaction analysis, in addition to step annotations, we also provide conversation transcripts and conversation-state annotations.

\subsection{Data Collection}

Our dataset includes videos in which an instructor teaches one of the four tasks (cart assembly, chair assembly, GoPro battery replacement, printer toner replacement) listed in Table \ref{tab:task-data} to a learner, who then executes the task. Ten learners (6 male, 4 female) and two instructors (1 male, 1 female) participated. The spatial layout of the instructor, the learner, and the cameras is shown in Fig. \ref{fig:KibanA-captured-position}. Before recording, instructors were given time to review the task procedures and, if necessary, consult an instruction sheet. To ensure visibility from both viewpoints, all tasks except printer toner replacement were performed face-to-face. Only the printer toner replacement was performed side by side. Each participant completed all four tasks, yielding 38 video sessions and a total of 8 hours of footage. Learners’ prior experience levels for each task are summarized in Fig. \ref{fig:skill-level}.

Both the instructor and the learner wore Aria Glasses \cite{ariaglasses} to record egocentric video. Through the Aria API, we can obtain gaze information and head pose from the camera trajectory, which we expect will be valuable for future interaction analysis. Additionally, we recorded two fixed third-person views with GoPro HERO11 cameras to enable multi-view analyses in future work, although these third-person views are not used in the experiments in this paper. All four cameras were time-synchronized using QR timecodes. Camera settings are listed in Table \ref{tab:camera-data}.

\subsection{Annotation}

\begin{figure*}[t]
  \centering
  \includegraphics[width=0.9\linewidth]{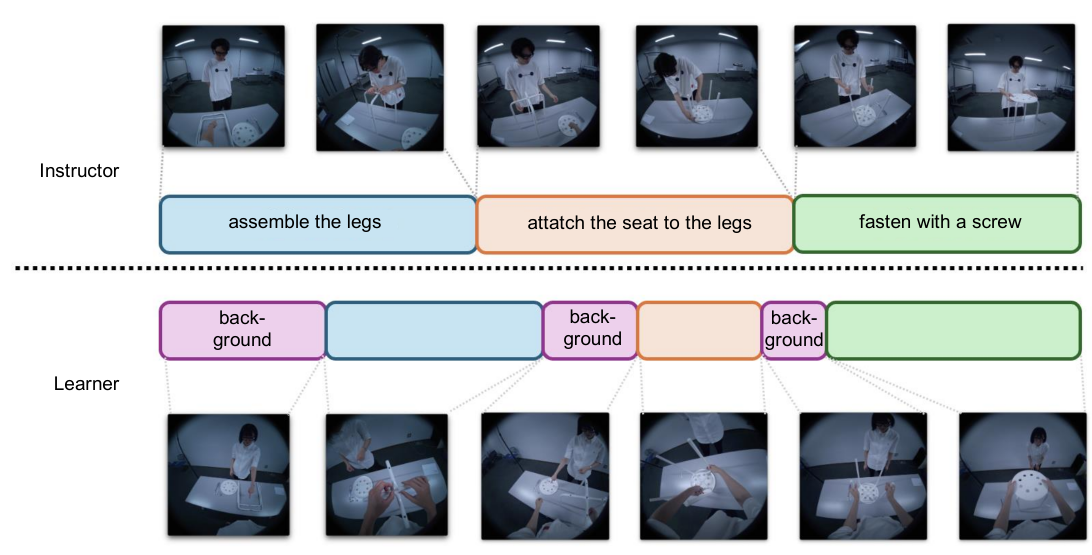}
  \caption{Examples of step annotation.}
  \label{fig:KibanA_annotation}
\end{figure*}

\begin{table}[t]
\centering
\caption{Tasks and steps in the EgoInstruct dataset.}
\setlength{\tabcolsep}{2mm}{
\begin{tabular}{@{}ll@{}}
  \toprule
   Task & Step \\
  \midrule
  Cart assembly & assemble two sidebars \\
   & assemble base frame \\
   & attach the sidebar \\
   & attach the lower tray \\
   & attach the middle tray \\
   & attach the upper tray \\
   & attach the four wheels \\
  Chair assembly &  assemble the legs\\
   & attatch the seat to the legs \\
   & fasten with a screw \\
  GoPro battery replacement & replace the battery \\
   & replace the microSD card\\
   & turn on power \\
   & turn off power \\
   & attach the peg \\
   & detach the peg \\
  Printer toner replacement & put paper \\
   & print in color \\
   & copy in black and white \\
   & replace cartridge \\
  \bottomrule
\end{tabular}
\label{tab:procedure-data}}
\end{table}

\begin{table*}[t]
  \centering
  \caption{Examples of utterances and their conversation-state labels. Utterances translated to English are shown.}
    \includegraphics[width=0.8\linewidth]{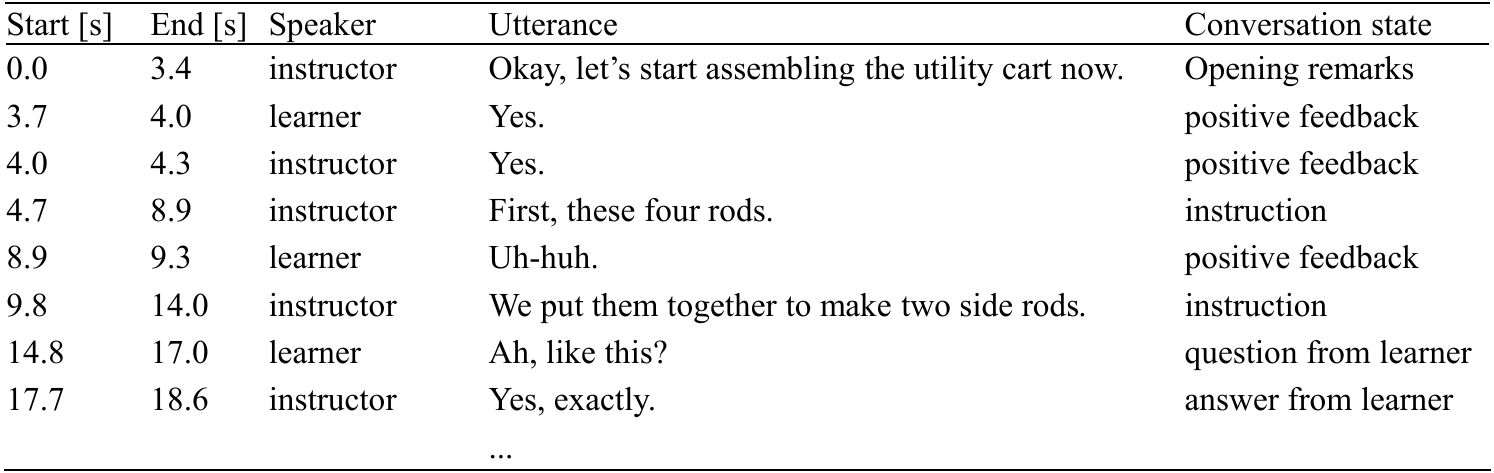}
  \label{tab:conversation-example}
\end{table*}

\begin{figure}[t]
    \centering
    \includegraphics[width=\linewidth]{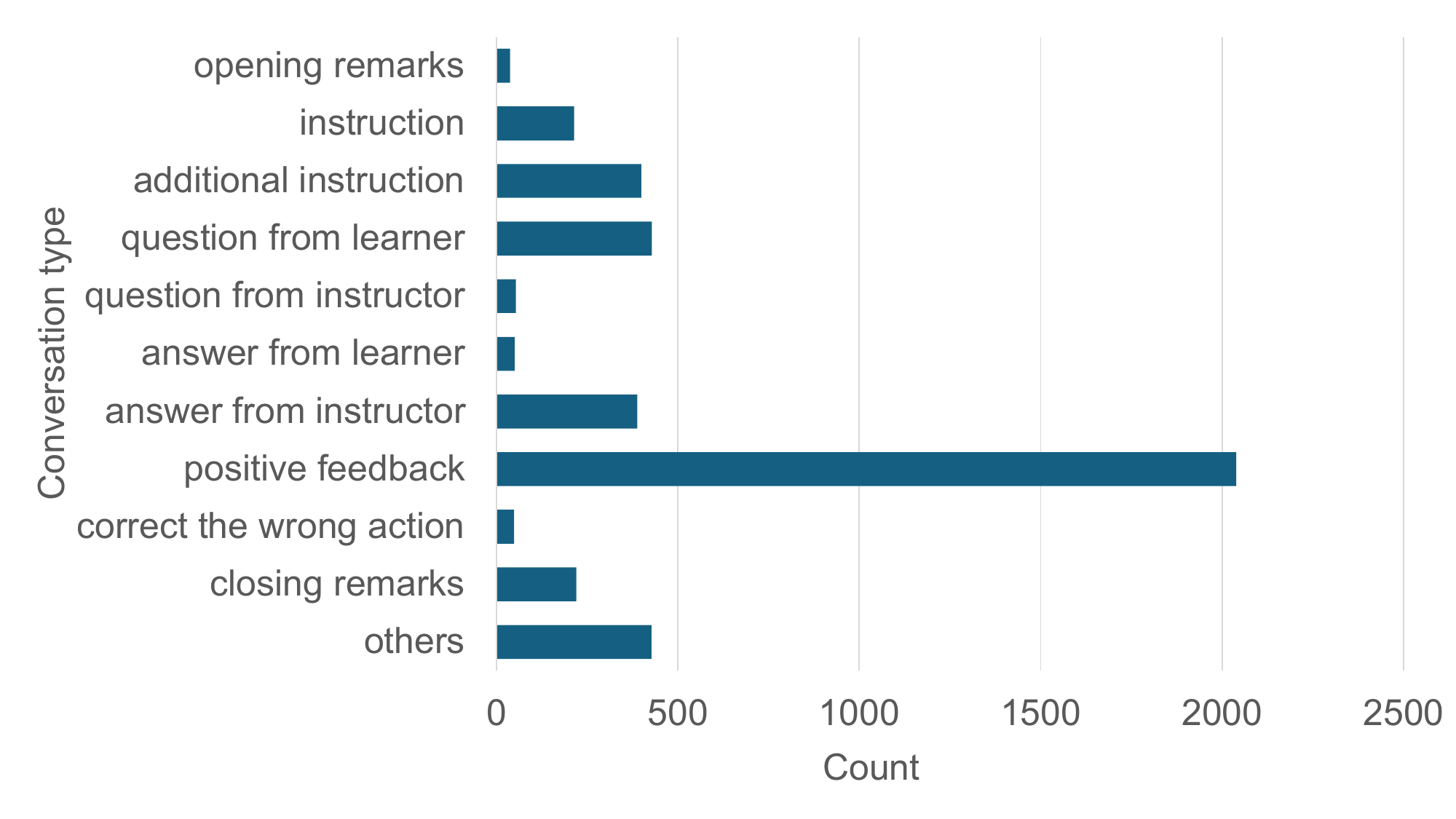}
    \caption{Distribution of conversation state labels.}
    \label{fig:KibanA-captured-overview-1}
\end{figure}

\noindent\textbf{Step Labels}
The four tasks in our dataset follow HoloAssist \cite{holoassist}, anticipating future use of HoloAssist data in pretraining or augmentation. Since the original tasks have different step granularities, we redefined step labels so that the granularity is consistent across all four tasks. The step definitions are provided in Table \ref{tab:procedure-data}.
  
We annotated steps separately for the instructor and the learner because the timing differs by viewpoint. For the instructor, a step starts when the instructor begins explaining it. For the learner, a step starts when the learner begins executing the step after listening to the instructor. We therefore provide distinct timestamps for steps from both viewpoints (Fig. \ref{fig:KibanA_annotation}).

\noindent\textbf{Utterances and Conversation States}
We transcribed the utterances of both the instructor and the learner and added timestamps because utterance timing is crucial for interaction analysis. Conversations were conducted in participants’ native language and transcribed in that language. Based on Porfirio et al.’s template for conversational state transitions in instructional scenes \cite{interaction-templete} and Wang et al.’s definitions \cite{holoassist}, we defined a set of conversation-state labels. Table \ref{tab:conversation-example} shows examples of annotated utterances, and Fig. \ref{fig:KibanA-captured-overview-1} summarizes the conversation-state classes and their distribution.

\section{Experiments}

We evaluated two tasks on our dataset: step segmentation from the learner’s viewpoint and conversation-state classification for utterances exchanged between the instructor and the learner.

\subsection{Step Segmentation}

\begin{table*}[t]
  \centering
  \caption{Step segmentation results on EgoInstruct across input modalities.}
  \setlength{\tabcolsep}{2mm}{
  \begin{tabular}{@{}lcccccccc@{}}
    \toprule
     Model & Video & Audio & Text  & F1$@$10 & F1$@$25 & F1$@$50 & Edit & Acc. \\
    \midrule
    ASFormer & \checkmark & & & 55.0 & 50.3 & 41.9 & 42.5 & 79.5\\ 
    GPT-4o & & & \checkmark& 80.7 & 72.3 & 58.9 & 86.6 & 58.8\\
    Gemini-2.5-pro & & & \checkmark & 94.0 & 91.1 & 85.5 & 89.6 & 83.1\\ 
    Gemini-2.5-pro & & \checkmark & & 91.6 & 87.9 & 73.1 & 89.7 & 75.9\\
    Gemini-2.5-pro & \checkmark & & & 81.8 & 73.5 & 55.2 & 81.6 & 62.8\\ 
    Gemini-2.5-pro & \checkmark & \checkmark & & 89.2 & 86.5 & 72.1 & 88.6 & 77.0\\
    Gemini-2.5-pro & \checkmark & \checkmark & \checkmark & {\bf 95.6} & {\bf 92.5} & {\bf 85.6} & {\bf 92.3} & {\bf 85.3}\\ 
    \bottomrule
  \end{tabular}
  \label{tab:video_segmentation}}
\end{table*}

\begin{figure*}[t]
    \centering
    \includegraphics[width=0.73\linewidth]{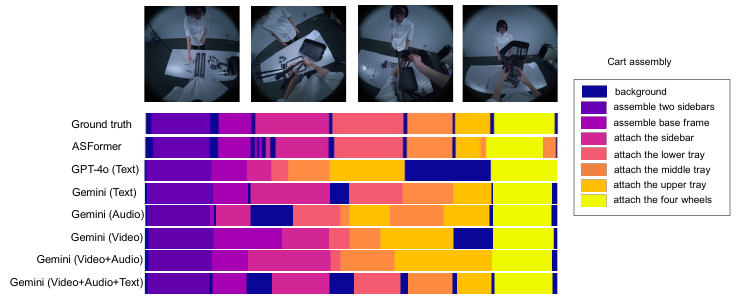}
    \includegraphics[width=0.73\linewidth]{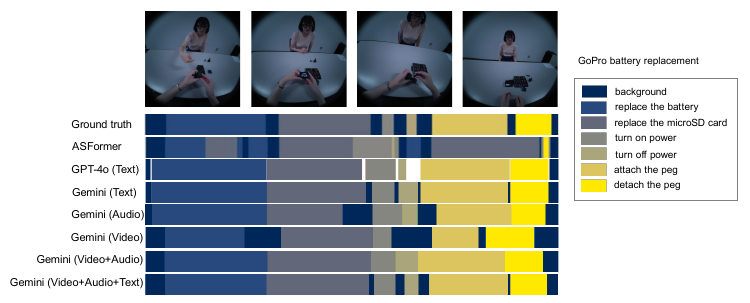}
    
    \caption{Examples of step segmentation results.}
    \label{KibanA_asformer_result1}
\end{figure*}

For step segmentation, we followed the standard action segmentation protocol, predicting the temporal intervals of procedural steps for each task. We compared ASFormer \cite{asformer}, a state-of-the-art method for action segmentation, against two MLLMs, Gemini-2.5-pro and GPT-4o.

\noindent\textbf{ASFormer}
We assessed how well step segmentation could be predicted from egocentric video when training on a subset of the dataset. We performed 5-fold cross-validation: in each fold, we trained on eight participants and tested on the remaining two. Only the learner’s egocentric video was used; third-person views were excluded. We trained with an initial learning rate of $5\times10^{-4}$ using the Adam optimizer for 20 epochs on a single NVIDIA A100-SXM4-40GB GPU.
We used I3D features extracted with a 3D ResNet backbone \cite{3D-resnet,i3d-resnet} pretrained on Kinetics \cite{i3d-feature}.

\noindent\textbf{MLLM Setup}
For Gemini-2.5-pro, we evaluated multiple input modalities: video only, audio only, text only, video+audio, and video+audio+text. We used the same test set as ASFormer to ensure a fair comparison. Gemini-2.5-pro was evaluated in a zero-shot setting with no training data. For a model comparison, we also evaluated GPT-4o under the text-only condition with the same test data. The prompts are provided in the supplementary material. The set of step labels was assumed known in the prompts, and the model was asked to output the start and end times of each step along with the label.

\noindent\textbf{Quantitative Results}
Table \ref{tab:video_segmentation} summarizes the results. With video-only input, zero-shot Gemini outperformed the trained ASFormer on F1 and Edit metrics, although ASFormer achieved higher frame-wise accuracy due to over-segmentation. When both models were evaluated under the same input modality (text only), Gemini-2.5-pro outperformed GPT-4o. Its performance was strong across individual modalities and improved further when modalities were combined; the best results were obtained with video+audio+text. These findings indicate that MLLMs effectively integrated multimodal cues to segment steps in face-to-face instruction.

\noindent\textbf{Qualitative Results}
Figure \ref{KibanA_asformer_result1} visualizes example segmentations. Compared to MLLMs, ASFormer tended to over-segment steps. We hypothesize that ASFormer primarily relied on visual pattern matching to actions seen during training, whereas MLLMs leveraged broad prior knowledge acquired during pretraining to produce more reasonable segmentations. We also observed that boundaries became more accurate when combining modalities. Purely visual input often lacks sufficient cues, and utterances that indicate the next step are not synchronized with the exact timing of the physical action. Multimodal fusion was therefore necessary for accurate boundaries.

\subsection{Conversation-State Classification}

\begin{table}[t]
  \centering
  \caption{Performance comparison of conversation type classification.}
  \setlength{\tabcolsep}{2mm}{
  \begin{tabular}{@{}lcccc@{}}
    \toprule
       & Acc & Precision & Recall & F1 score \\
    \midrule
    Without context & & & \\
    BERT & {\bf 0.76} & {\bf 0.76} & {\bf 0.76} & {\bf 0.74}\\ 
    GPT-4o & 0.72 & 0.73 & 0.72 & 0.71\\ 
    Gemini-2.5-pro & 0.74 & 0.74 & 0.74 & 0.72\\ 
    \midrule
    With context & & & \\
    GPT-4o & 0.78 & 0.81 & 0.78 & 0.77\\ 
    Gemini-2.5-pro & {\bf 0.84} & {\bf 0.86} & {\bf 0.84} & {\bf 0.84}\\ 
    \bottomrule
  \end{tabular}
  \label{tab:KibanA-conversation-result}}
\end{table}

\begin{figure*}[t]
    \centering
    \includegraphics[width=\linewidth]{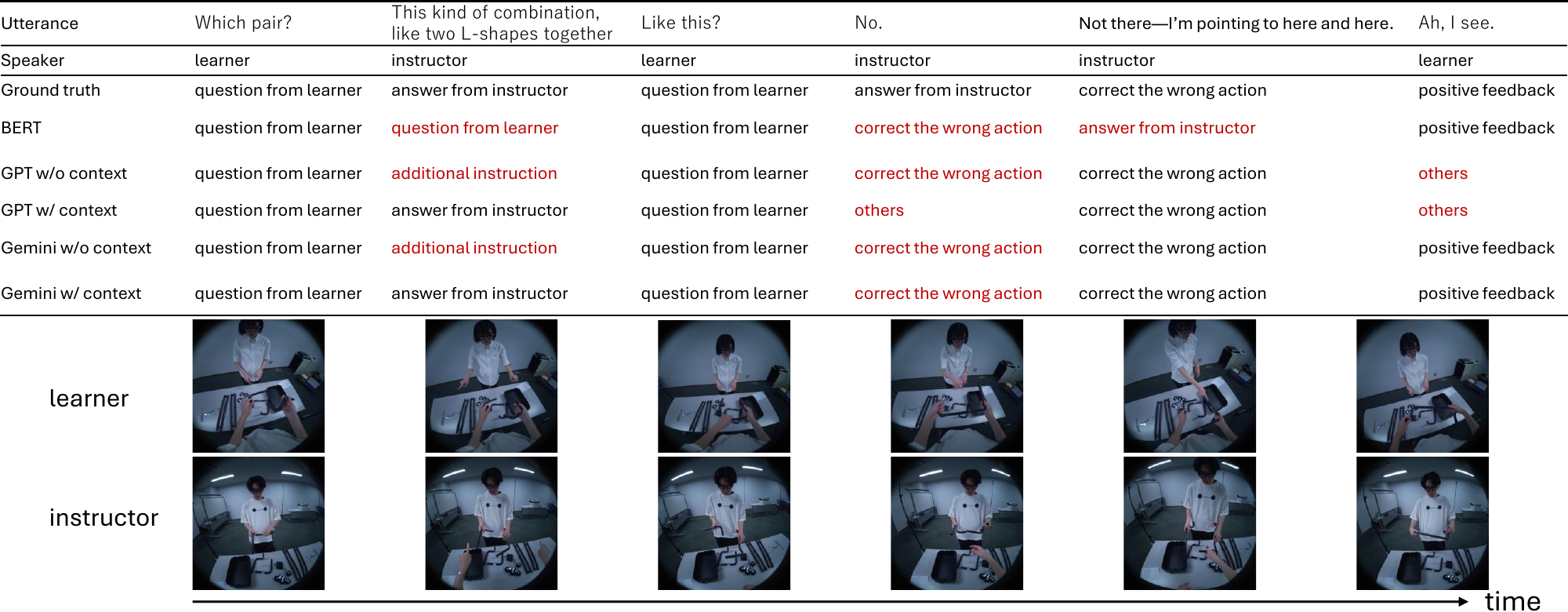}
    \caption{Conversation-state classification results. Utterances translated to English are shown.}
    \label{fig:KibanA_result_conversation}
\end{figure*}

As a first step toward comprehensive understanding of face-to-face instruction, we classified the conversation state for each utterance. All utterances in the dataset are labeled with conversation-state annotations. We took the utterance text as input and predicted the conversation-state label. We compared a widely used NLP baseline, BERT \cite{bert}, with the LLMs Gemini-2.5-pro and GPT-4o.

\noindent\textbf{BERT}
We used a pre-trained multilingual BERT with 24 Transformer layers, a hidden size of 1024, and 16 attention heads. We trained with AdamW at an initial learning rate of $5\times10^{-4}$ for 20 epochs on a single NVIDIA A100-SXM4-40GB GPU. We performed 5-fold cross-validation: eight participants for training and two for testing in each fold. Each utterance (one sentence) was treated as an individual input.

\noindent\textbf{LLM}
GPT-4o and Gemini-2.5-pro were used without additional fine-tuning. Following the in-context learning paradigm, we supplied in-context exemplars from eight participants, namely pairs of utterances with conversation-state labels, and performed inference on the remaining two participants’ test set. We evaluated two settings: \emph{without context}, where each utterance was predicted independently as in BERT, and \emph{with context}, where we provided the entire conversation sequence in the test set and the model predicted all utterances jointly with access to surrounding context. Prompts are given in the supplementary material.

\noindent\textbf{Quantitative Results}
Table \ref{tab:KibanA-conversation-result} shows the results. Without context, BERT achieved higher performance. This can be interpreted as a trade-off between BERT’s task specialization and the LLMs’ broader world knowledge. With fine-tuning on a specific dataset, BERT could replicate task-specific labeling rules precisely, which was advantageous when clear criteria and adherence to given examples were more important than outside knowledge. LLMs possess strong internal world models from pretraining. When this prior knowledge conflicted with task-specific rules, the model sometimes favored its own ``common sense,'' leading to reasonable but task-inconsistent labels and thus lower accuracy.

\noindent\textbf{Qualitative Results}
Overall, labels corresponding to instructional speech, such as \emph{instruction} and \emph{additional instruction}, were often predicted correctly, likely because these utterances were longer and contained enough information within a single sentence. Figure \ref{fig:KibanA_result_conversation} shows qualitative examples. Without context, we observed errors such as answering before the question was asked or swapping speaker roles, but these were largely resolved when the full conversational context was provided.

\section{Conclusion}

We presented a new egocentric dataset of face-to-face instruction and analyzed instructional interactions using it. Existing datasets rarely focus on instruction, making interaction analysis in instructional scenes difficult. Our dataset provides egocentric views from both the instructor and the learner along with annotations for conversation states and procedural steps. We evaluated models, including multimodal large language models (MLLMs), on two tasks: procedural step segmentation and conversation-state classification. For step segmentation, MLLMs produced accurate boundaries by leveraging multiple modalities. For conversation-state classification, LLMs achieved strong performance when provided with conversational context. These results suggest that MLLMs can contribute to holistic understanding of face-to-face instruction even without additional fine-tuning.

Future work includes scaling up the dataset and tackling more advanced tasks for understanding instructional scenes. As a first step, we built a dataset and focused on fundamental tasks in this paper. Examples of more advanced tasks include recognizing deictic expressions such as “this” and “here,” estimating the learner’s level of understanding, and extracting tips and tacit knowledge that are not written in manuals; we will then evaluate the extent to which MLLMs can advance on these fronts. Another direction is to exploit information we recorded but did not use in this paper (third-person views, gaze, and SLAM-based 3D trajectories) and to devise effective ways to incorporate these signals into MLLMs to enable deeper understanding.

\noindent\textbf{Acknowledgement}
This work was supported by JST ASPIRE Grant Number JPMJAP2303, JSPS KAKENHI Grant Number 23H00488, and 24K02956.

\clearpage

\section{Other Results}

\subsection{Step Segmentation}

Table \ref{tab:video_segmentation_each_task} shows the step segmentation results on each task: cart assembly, chair assembly, GoPro battery replacement, and printer toner replacement.

\begin{table*}[t]
  \centering
  \caption{Step segmentation results on each task.}
  \setlength{\tabcolsep}{2mm}{
  \begin{subtable}{\textwidth}
      \centering
      \caption{Cart assembly}
      \begin{tabular}{@{}lcccccccc@{}}
        \toprule
         Model & Video & Audio & Text  & F1$@$10 & F1$@$25 & F1$@$50 & Edit & Acc. \\
        \midrule
        ASFormer & \checkmark & & & 68.6 & 62.9 & 57.1 & 53.1 & {\bf 91.6}\\ 
        GPT-4o & & & \checkmark& 72.9 & 62.9 & 46.8 & 82.0 & 49.6\\
        Gemini-2.5-pro & & & \checkmark & 97.8 & 96.3 & 94.7 & 95.9 & 85.6\\ 
        Gemini-2.5-pro & & \checkmark & & 89.4 & 88.0 & 83.8 & 93.1 & 79.0\\
        Gemini-2.5-pro & \checkmark & & & 75.0 & 70.9 & 52.7 & 84.8 & 58.9\\ 
        Gemini-2.5-pro & \checkmark & \checkmark & & 84.4 & 83.0 & 70.6 & 86.3 & 72.5\\
        Gemini-2.5-pro & \checkmark & \checkmark & \checkmark & {\bf 99.3} & {\bf 97.9} & {\bf 97.9} & {\bf 98.8} & 88.1\\ 
        \bottomrule
      \end{tabular}
  \end{subtable}

  \begin{subtable}{\textwidth}
      \centering
      \caption{Chair assembly}
      \begin{tabular}{@{}lcccccccc@{}}
        \toprule
         Model & Video & Audio & Text  & F1$@$10 & F1$@$25 & F1$@$50 & Edit & Acc. \\
        \midrule
        ASFormer & \checkmark & & & 66.7 & 66.7 & {\bf 66.7} & 56.3 & {\bf 95.8}\\ 
        GPT-4o & & & \checkmark& 64.0 & 54.5 & 38.3 & 69.5 & 51.3\\
        Gemini-2.5-pro & & & \checkmark & 80.7 & 70.7 & 56.7 & 67.5 & 77.3\\ 
        Gemini-2.5-pro & & \checkmark & & 80.4 & 75.0 & 53.6 & 71.5 & 75.1\\
        Gemini-2.5-pro & \checkmark & & & 71.8 & 55.3 & 30.6 & 60.8 & 64.2\\ 
        Gemini-2.5-pro & \checkmark & \checkmark & & 76.7 & 69.5 & 43.4 & 70.5 & 70.1\\
        Gemini-2.5-pro & \checkmark & \checkmark & \checkmark & {\bf 84.3} & {\bf 76.4} & 65.7 & {\bf 73.0} & 82.6\\ 
        \bottomrule
      \end{tabular}
  \end{subtable}

  \begin{subtable}{\textwidth}
      \centering
      \caption{GoPro battery replacement}
      \begin{tabular}{@{}lcccccccc@{}}
        \toprule
         Model & Video & Audio & Text  & F1$@$10 & F1$@$25 & F1$@$50 & Edit & Acc. \\
        \midrule
        ASFormer & \checkmark & & & 34.8 & 21.7 & 8.7 & 27.2 & 51.3\\ 
        GPT-4o & & & \checkmark& 98.3 & 98.3 & 90.0 & {\bf 100.0} & 82.0\\
        Gemini-2.5-pro & & & \checkmark & 98.5 & {\bf 98.5} & {\bf 93.7} & 97.1 & {\bf 87.4}\\ 
        Gemini-2.5-pro & & \checkmark & & 97.8 & 97.8 & 86.4 & 96.1 & 79.4\\
        Gemini-2.5-pro & \checkmark & & & 91.8 & 85.1 & 71.5 & 86.7 & 73.3\\ 
        Gemini-2.5-pro & \checkmark & \checkmark & & 95.5 & 95.5 & 82.4 & 97.5 & 81.9\\
        Gemini-2.5-pro & \checkmark & \checkmark & \checkmark & {\bf 98.6} & 95.5 & 88.8 & 97.5 & 85.1\\ 
        \bottomrule
      \end{tabular}
  \end{subtable}

  \begin{subtable}{\textwidth}
      \centering
      \caption{Printer toner replacement}
      \begin{tabular}{@{}lcccccccc@{}}
        \toprule
         Model & Video & Audio & Text  & F1$@$10 & F1$@$25 & F1$@$50 & Edit & Acc. \\
        \midrule
        ASFormer & \checkmark & & & 50.0 & 50.0 & 35.0 & 33.5 & 79.1\\ 
        GPT-4o & & & \checkmark& 87.4 & 73.4 & 60.3 & 94.7 & 52.1\\
        Gemini-2.5-pro & & & \checkmark & 98.9 & 98.9 & {\bf 96.7} & 98.0 & 82.2\\ 
        Gemini-2.5-pro & & \checkmark & & 98.9 & 90.9 & 68.4 & 98.0 & 70.0\\
        Gemini-2.5-pro & \checkmark & & & 88.7 & 82.7 & 65.8 & 94.0 & 54.7\\ 
        Gemini-2.5-pro & \checkmark & \checkmark & & {\bf 100.0} & 98.0 & 92.0 & {\bf 100.0} & 83.4\\
        Gemini-2.5-pro & \checkmark & \checkmark & \checkmark & {\bf 100.0} & {\bf 100.0} & 90.0 & {\bf 100.0} & {\bf 85.5}\\ 
        \bottomrule
      \end{tabular}
  \end{subtable}
  
  \label{tab:video_segmentation_each_task}}
\end{table*}

\begin{figure*}[t]
    \centering
    \includegraphics[width=0.73\linewidth]{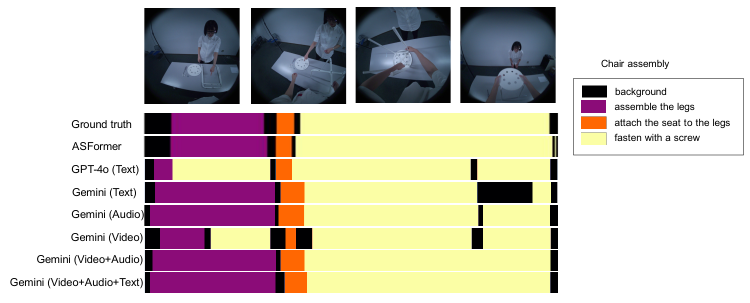}
    \includegraphics[width=0.73\linewidth]{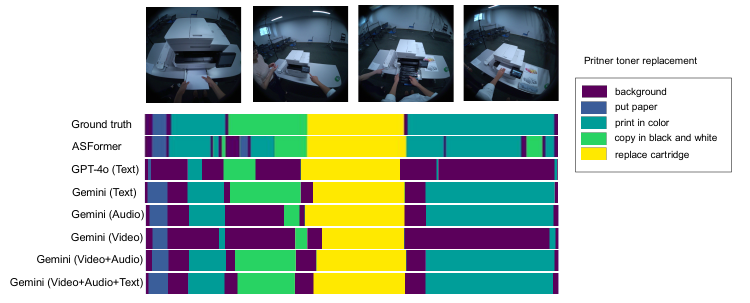}
    
    \caption{Examples of step segmentation results.}
    \label{KibanA_asformer_result2}
\end{figure*}

Figure \ref{KibanA_asformer_result2} visualizes the results of step segmentation on ``chair assembly'' and ``Printer toner replacement'' tasks.

\subsection{Conversation-state classification}

Table \ref{tab:KibanA-conversation-result_each_task} shows the conversation-state classification results on each task: cart assembly, chair assembly, GoPro battery replacement, and printer toner replacement.

\begin{table*}[t]
  \centering
  \caption{Performance comparison of conversation type classification on each task.}
  \setlength{\tabcolsep}{2mm}{
  \begin{subtable}{\textwidth}
      \centering
      \caption{Cart assembly}
      \begin{tabular}{@{}lcccc@{}}
        \toprule
           & Acc & Precision & Recall & F1 score \\
        \midrule
        Without context & & & \\
        BERT & {\bf 0.69} & {\bf 0.72} & {\bf 0.69} & {\bf 0.69}\\ 
        GPT-4o & 0.63 & 0.64 & 0.63 & 0.62\\ 
        Gemini-2.5-pro & 0.67 & 0.68 & 0.67 & 0.66\\ 
        \midrule
        With context & & & \\
        GPT-4o & 0.70 & 0.74 & 0.70 & 0.68\\ 
        Gemini-2.5-pro & {\bf 0.77} & {\bf 0.80} & {\bf 0.77} & {\bf 0.76}\\ 
        \bottomrule
      \end{tabular}
  \end{subtable}

  \begin{subtable}{\textwidth}
      \centering
      \caption{Chair assembly}
      \begin{tabular}{@{}lcccc@{}}
        \toprule
           & Acc & Precision & Recall & F1 score \\
        \midrule
        Without context & & & \\
        BERT & 0.70 & 0.68 & 0.70 & 0.66\\ 
        GPT-4o & 0.73 & 0.73 & 0.73 & 0.71\\ 
        Gemini-2.5-pro & {\bf 0.76} & {\bf 0.76} & {\bf 0.76} & {\bf 0.74}\\ 
        \midrule
        With context & & & \\
        GPT-4o & 0.81 & 0.82 & 0.81 & 0.79\\ 
        Gemini-2.5-pro & {\bf 0.87} & {\bf 0.89} & {\bf 0.87} & {\bf 0.87}\\ 
        \bottomrule
      \end{tabular}
  \end{subtable}

  \begin{subtable}{\textwidth}
      \centering
      \caption{GoPro battery replacement}
      \begin{tabular}{@{}lcccc@{}}
        \toprule
           & Acc & Precision & Recall & F1 score \\
        \midrule
        Without context & & & \\
        BERT & {\bf 0.82} & {\bf 0.83} & {\bf 0.82} & {\bf 0.79}\\ 
        GPT-4o & 0.73 & 0.75 & 0.73 & 0.72\\ 
        Gemini-2.5-pro & 0.75 & 0.75 & 0.75 & 0.73\\ 
        \midrule
        With context & & & \\
        GPT-4o & 0.81 & 0.83 & 0.81 & 0.80\\ 
        Gemini-2.5-pro & {\bf 0.85} & {\bf 0.87} & {\bf 0.85} & {\bf 0.85}\\ 
        \bottomrule
      \end{tabular}
  \end{subtable}

  \begin{subtable}{\textwidth}
      \centering
      \caption{Printer toner replacement}
      \begin{tabular}{@{}lcccc@{}}
        \toprule
           & Acc & Precision & Recall & F1 score \\
        \midrule
        Without context & & & \\
        BERT & {\bf 0.82} & {\bf 0.80} & {\bf 0.82} & {\bf 0.80}\\ 
        GPT-4o & 0.79 & 0.79 & 0.79 & 0.78\\ 
        Gemini-2.5-pro & 0.78 & 0.78 & 0.78 & 0.77\\ 
        \midrule
        With context & & & \\
        GPT-4o & 0.83 & 0.84 & 0.83 & 0.82\\ 
        Gemini-2.5-pro & {\bf 0.87} & {\bf 0.88} & {\bf 0.87} & {\bf 0.87}\\ 
        \bottomrule
      \end{tabular}
  \end{subtable}
  \label{tab:KibanA-conversation-result_each_task}}
\end{table*}

\section{MLLM Prompt}

\subsection{Step Segmentation}
Figure~\ref{fig:prompt_step} shows the prompt used for procedural step segmentation.

\begin{figure*}[t]
\begin{promptbox}
    This is a video of an instructor teaching a student how to use or assemble something. You are an expert AI assistant specializing in analyzing and segmenting procedural tasks. Your goal is to segment the provided media into a sequence of high-level, distinct procedural steps.
    
    Your Instructions:
    \begin{promptlist}
        \item Identify High-Level Tasks: Analyze the user's actions and conversation to identify the major stages of the assembly process.
        \item Use the Provided Labels: You must use only the following labels for the segments: assemble\_two\_sidebars, assemble\_base\_frame, attach\_the\_sidebar, attach\_the\_upper\_tray, attach\_the\_middle\_tray, attach\_the\_lower\_tray, attach\_the\_four\_wheels, and background
        \item Apply Segmentation Rules:
        \begin{promptenum}
            \item Task Boundaries: A task segment should start when the physical action of that task begins (e.g., picking up the relevant parts) and end when the action for that specific task is fully completed (e.g., all screws are tightened, a part is declared finished). You will be evaluated on how precise you predicted the boundaries. When multiple media are provided, prioritize video and visual information over text and audio.
            \item background Label: This is a special label. Use it for any period that is not a primary assembly task. This includes:
            \begin{promptlist}
                \item The initial opening remarks and greetings before the first action.
                \item Transitions between major assembly steps where participants are pausing, discussing the next step, or reorienting themselves without actively manipulating the core parts of the current task.
                \item The final closing remarks after the assembly is complete.
            \end{promptlist}
            \item Capture the most subtle nuances of the segment labels, and be sure that every single element of the label is fulfilled before assigning.
        \end{promptenum}
    \end{promptlist}
    Output Format:
    \begin{promptlist}
        \item The output must be a single JSON array of objects.
        \item Each object must have three keys: "start", "end", and "text". Write time in seconds format to the second decimal (not in minutes+seconds format), and write the label in "text"
        \item The segmentation must be continuous, covering the entire duration of the transcript. The end time of one segment must be the start time of the next.
    \end{promptlist}
    Analyze the following transcript and generate the JSON output based on these rules. Don't return anything other than the JSON output. Don't show '''JSON either.
  \end{promptbox}
  \caption{Input prompt for procedural step segmentation.}
  \label{fig:prompt_step}
\end{figure*}

\subsection{Conversation-State Classification}

Figure \ref{fig:prompt_conversation} shows the prompt used for conversation-state classification.

\begin{figure*}[t]
\begin{promptbox}
    You are a Japanese dialogue-act labeler. Each dialogue must be labeled exactly one of:
    \begin{promptenum}
        \item opening remarks
        \item instruction
        \item additional instruction
        \item question from learner
        \item question from instructor
        \item answer from learner
        \item answer from instructor
        \item correct the wrong action
        \item positive feedback
        \item closing remarks
        \item others
    \end{promptenum}
    Return **only** a valid JSON object in the format: {{"id": 0, "label": "predicted\_label"}}. Do not include any additional text, explanations, or code block markers (e.g., ```json). If multiple categories seem possible, choose the one that best fits the main function of the utterance. Use the following training data to refine your predictions.

    [label0]: [training\_utterance\_text0]
    
    [label1]: [trainig\_utterance\_text1]\\
    ...
    
    Now, predict the label for the following utterance:

    id: 0, [test\_utterance\_text0]

    id: 1, [test\_utterance\_text1]\\
    ...
  \end{promptbox}
  \caption{Input prompt for conversation-state classification.}
  \label{fig:prompt_conversation}
\end{figure*}

\end{document}